\documentclass{article}
\usepackage{arxiv}
\usepackage{xcolor}
\usepackage[utf8]{inputenc} 
\usepackage[T1]{fontenc}    
\usepackage{hyperref}       
\usepackage{url}            
\usepackage{booktabs}       
\usepackage{amsfonts}       
\usepackage{nicefrac}       
\usepackage{microtype}      
\usepackage{lipsum}
\usepackage{graphicx}
\usepackage{amsmath}
\usepackage{amssymb} 
\usepackage{dsfont}
\usepackage{graphicx}
\usepackage{algorithm}
\usepackage{algorithmic}
\newcommand{\bT}{\mathbf{T}}
\newcommand{\bA}{\mathbf{A}}
\newcommand{\bX}{\mathbf{X}}
\newcommand{\bY}{\mathbf{Y}}
\newcommand{\bG}{\mathbf{G}}
\newcommand{\bZ}{\mathbf{Z}}
\newcommand{\bW}{\mathbf{W}}
\newcommand{\bF}{\mathbf{F}}
\newcommand{\bI}{\mathbf{I}}
\newcommand{\bS}{\mathbf{S}}
\newcommand{\bD}{\mathbf{D}}
\newcommand{\bx}{\mathbf{x}}

\usepackage{tabularx}
\usepackage{adjustbox}

\title{Graph-Convolutional Networks: Named Entity
Recognition and Large Language Model
Embedding in Document Clustering}

\author{
 Imed Keraghel \\
  Centre Borelli UMR9010\\
  Université Paris Cité,\\
  Paris, France \\
  \texttt{imed.keraghel@u-paris.fr} \\
   \And
 Mohamed Nadif \\
  Centre Borelli UMR9010\\
  Université Paris Cité,\\
  Paris, France \\
  \texttt{mohamed.nadif@u-paris.fr} \\
  } 

\date{}

\begin{document}
\maketitle
\begin{abstract}
Recent advances in machine learning, particularly Large Language Models (LLMs) such as BERT and GPT, provide rich contextual embeddings that improve text representation. However, current document clustering approaches often ignore the deeper relationships between named entities (NEs) and the potential of LLM embeddings. This paper proposes a novel approach that integrates Named Entity Recognition (NER) and LLM embeddings within a graph-based framework for document clustering. The method builds a graph with nodes representing documents and edges weighted by named entity similarity, optimized using a graph-convolutional network (GCN). This ensures a more effective grouping of semantically related documents. Experimental results indicate that our approach outperforms conventional co-occurrence-based methods in clustering, notably for documents rich in named entities. 
\end{abstract}


\section{Introduction}

Document clustering is widely used in data analytics, especially in fields like information retrieval and natural language processing (NLP). It groups documents into categories based on shared characteristics, which is crucial for tasks like topic modeling and recommendation systems. Traditional methods depend on lexical features \cite{schutze2008introduction,wei2015semantic}, co-occurrences, or TF-IDF matrices \cite{affeldt2021regularized,salah2019directional}. Although effective in various situations, these methods face challenges in recognizing semantic connections that extend beyond basic word frequency analysis \cite{keraghel2024beyond}.

Recent advances in machine learning, particularly with the emergence of Large Language Models (LLMs) such as BERT \cite{vaswani2017attention,lee2018pre} and GPT \cite{radford2018improving}, have revolutionized text representation. These models offer rich contextual embeddings that capture the nuances of word meanings in context. However, many clustering methods still use traditional approaches like k-Nearest Neighbors (KNN) to construct graphs \cite{kim2021knn,qin2018novel}, which depend on shallow lexical similarities and often fail to capture deeper semantic relationships between documents.

To illustrate the limitations of traditional methods, Figure \ref{fig.graph.ner} compares two graph structures generated from the BBC News dataset: one using a KNN-based graph and the other using our proposed NER-based graph construction method. In the KNN graph, documents are connected based on lexical similarity, leading to overlapping and indistinct clusters. In contrast, the NER-based graph leverages named entity similarities, resulting in a clearer separation of clusters that correspond to distinct semantic topics.
\begin{figure}[!ht] 
\centering \includegraphics[width=11cm]{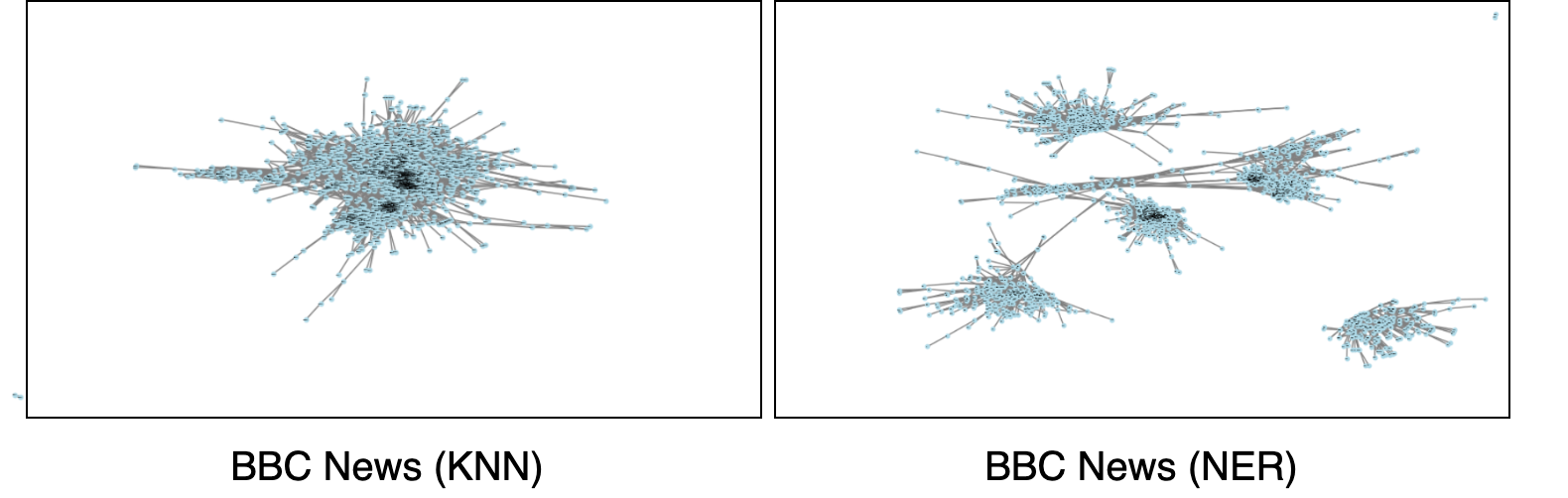} \caption{Comparison of graph structures for the BBC News dataset. (Left) KNN-based graph with lexical similarity; (Right) NER-based graph capturing entity similarities.} 
\label{fig.graph.ner} 
\end{figure}

In this paper, we propose a novel approach that leverages Named Entity Recognition (NER) alongside LLM embeddings in a graph-based framework for document clustering. Our method builds a graph where nodes represent documents and edges are weighted by the similarity of named entity contexts in each document. Named entities in similar contexts show strong semantic connections, and combining NER with LLM embeddings constructs a graph that better represents document similarities. To optimize this, we use a Graph Convolutional Network (GCN) \cite{fettal2022efficient,kipf2016semi}, which enables the joint optimization of the embeddings and the clustering objectives. This procedure guarantees that semantically related documents are grouped more effectively, addressing the shortcomings of conventional clustering techniques that do not incorporate global contextual information. Our main contributions include:
\begin{itemize} 
\item A document clustering method that improves accuracy using NER, LLM embeddings, and graph-based representations.
\item A technique to create an adjacency matrix based on named entity similarity for precise document relationship identification.
\item Experiments demonstrating our approach outperforms co-occurrence-based techniques, especially for documents rich in named entities. 
\end{itemize}

The remainder of this paper is organized as follows. Section 2 reviews the literature on graph representation learning, graph clustering, GCNs, LLMs, and NER. Section 3 details our method, including adjacency matrix construction and GCNs. Section 4 presents the experimental framework and results, followed by a conclusion and future perspectives in Section 5.

\section{Related work}
Unsupervised graph representation learning has seen remarkable progress in recent years, primarily through two key approaches: contrastive learning and autoencoders. Contrastive learning has emerged as a powerful method due to its ability to differentiate between positive and negative pairs in a self-supervised manner. Techniques such as GraphCL \cite{you2020graph} have introduced graph augmentations that improve graph representations by maximizing the agreement between different augmented views of the same graph. Similarly, MVGRL \cite{hassani2020contrastive} enhances the embeddings of node- and graph-level by contrasting views at both levels. On the other hand, autoencoder-based methods provide an alternative by reconstructing specific graph properties, such as the adjacency matrix, to learn embeddings. For example, Graph Autoencoders (GAEs), introduced by \cite{wang2016structural}, propose an architecture designed for large-scale graph data, enabling tasks such as node classification and link prediction. Further developments, such as Variational Graph Autoencoders (VGAE) \cite{kipf2016variational}, extend this approach by incorporating variational autoencoders with GCNs, while Linear Variational Graph Autoencoders (LVGAE) \cite{salha2021simple} simplify this architecture using a linear transformation with a one-hop propagation matrix.

Building on the foundation of graph representation learning, graph clustering aims to group nodes into clusters based on either graph structure or node-level features, or both. When focusing solely on node attributes, traditional clustering algorithms can be applied to graph data. Spectral clustering \cite{ng2001spectral}, for example, leverages the graph structure exclusively and optimizes the ratio and normalized cut criteria to form clusters. Deep Graph Infomax (DGI) \cite{velickovic2019deep}, on the other hand, improves graph clustering by maximizing mutual information between graph embeddings and substructures. In addition to these approaches, more recent methods have explored joint optimization of node embeddings and clustering objectives. An example is Graph Convolutional Clustering (GCC) \cite{fettal2022efficient}, which simultaneously learns node embeddings and clusters, resulting in enhanced performance and reduced computational costs. These methods show that incorporating clustering into the learning process can result in more meaningful representations and improved clustering outcomes.

In parallel to advancements in graph learning, recent progress in LLMs has significantly improved document representation. Models such as BERT or GPT provide deep contextual embeddings that are particularly well suited for tasks such as clustering, as they capture semantic relationships between entities and documents that traditional methods often miss \cite{muennighoff2022mteb,keraghel2024beyond}. Sentence-BERT \cite{reimers2019sentence}, for example, fine-tunes BERT to specialize in sentence similarity, making it an ideal candidate for clustering semantically related texts. Incorporating clustering algorithms such as k-means directly into LLM embeddings has also been explored, as seen with BERT-Kmeans \cite{sia2020tired,keraghel2024beyond}. Furthermore, the combination of LLMs with graph-based models, such as Graph-BERT \cite{zhang2020graph}, provides a promising avenue for more accurate and context-aware document clustering by integrating semantic knowledge and graph structures.

\section{Model and algorithm}
\begin{figure}[ht!]
  \centering
  \fbox{\includegraphics[width=13cm]{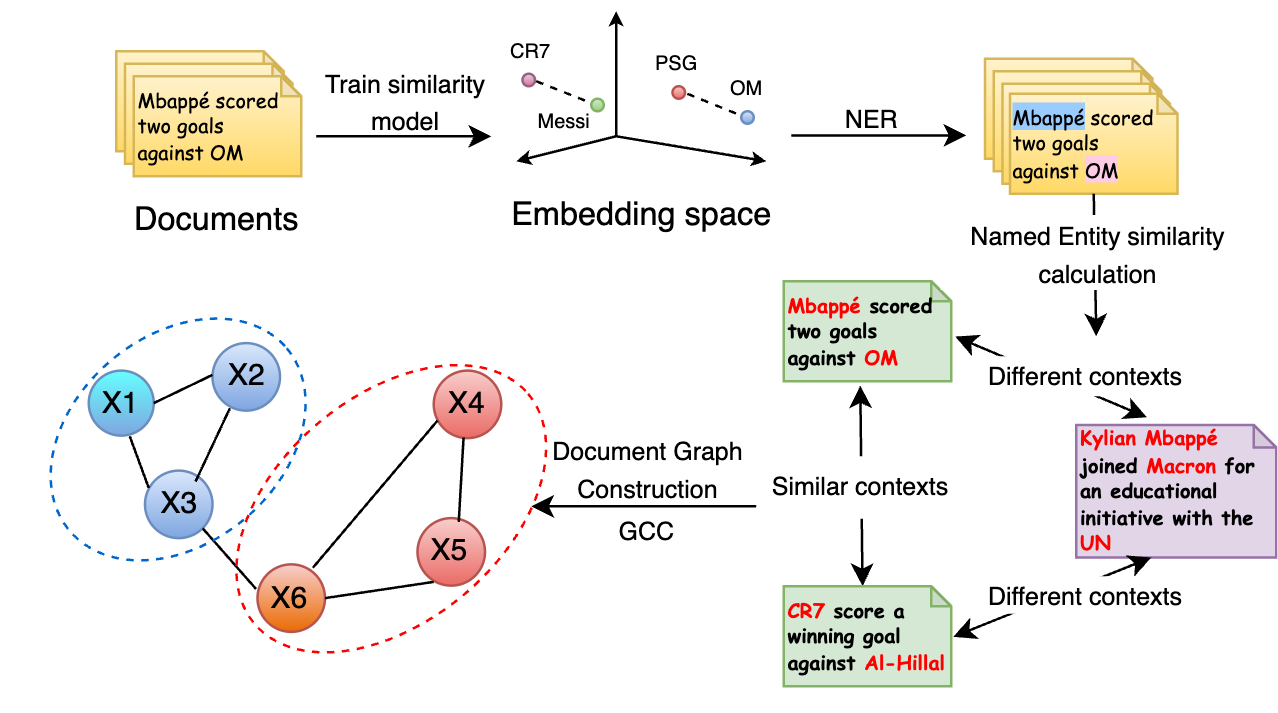}}
  \caption{Overview of the proposed model pipeline: LLM-based feature extraction, NER-based graph construction, and joint embedding and clustering.}
  \label{fig1}
\end{figure}

The use of NEs in clustering, although not yet widely adopted, presents a promising approach to improve document clustering. By focusing on key entities, such as medical concepts, NE-based clustering can group documents that share relevant entities. For example, in sports news, named entities such as "Kylian Mbappé" and "Cristiano Ronaldo" can link documents about their performances, such as a match in which Mbappé leads PSG to victory or Ronaldo secures Portugal's World Cup semi-final spot. Furthermore, their involvement in political campaigns, like a UN-backed global education initiative alongside "Emmanuel Macron," can connect documents from political domain. Research such as \cite{cao2012text,derczynski2015analysis} shows that the integration of NEs into clustering can lead to improved performance, especially in fields where entities are central to the thematic structure of the text.
We start by presenting the preliminaries and notation, followed by a description of our proposed model and algorithm (Figure \ref{fig1}).
\subsection{Preliminaries and Notation}

Let $\cal{G}= (V, \bA, \bX)$ represent an undirected graph, where: $\cal{V}$ is the set of vertex, consisting of nodes $\{ v_1, \dots, v_n \}$,
$\bA \in \mathbb{R}^{n \times n}$ is a symmetric adjacency matrix, where $a_{ij}$ indicates the edge weight between nodes $v_i$ and $v_j$,
$\bX \in \mathbb{R}^{n \times d}$ is the matrix of node features. In the following, we adopt the following notations:
$\text{Tr}(\cdot)$ denotes the trace of a matrix,
$k$ is the number of clusters,
$e$ is the embedding dimension,
$\mathds{1}_m$ is a column vector of $m$ ones, and
$\bI_m$ is the identity matrix of dimension $m$. For matrix $\bX$, $\bx_i$ refers to the $i$-th row vector, $\bx'_j$ to the $j$-th column vector, and $x_{ij}$ to the element in the $i$-th row and $j$-th column.

\subsection{Leveraging LLM Embeddings to Represent Node Features}
In our approach, the node feature matrix $\bX \in \mathbb{R}^{n \times d}$, which typically represents document features, is now replaced by LLM embeddings and $d$ becomes ${d_{\ell \ell m}}$. Traditional feature representations such as TF-IDF focus on surface-level word frequencies and co-occurrence patterns, which may not capture semantic nuances. Using LLM embeddings, we aim to incorporate rich contextual information that provides a more holistic view of the document semantics.

Let $f_{\ell\ell m}: {\cal{D}} \to \mathbb{R}^{d_{\ell\ell m}}$ be the LLM model that maps each document $d_i \in \cal{D}$ or its components (words, sentences, or entities) to a vector in a high-dimensional embedding space. The features matrix $X_{\ell \ell m}$ is constructed as:
\begin{equation} 
\bX_{\ell \ell m} = \begin{bmatrix}
    f_{\ell\ell m}(d_1) \\
    f_{\ell\ell m}(d_2) \\
    \vdots \\
    f_{\ell\ell m}(d_n)
\end{bmatrix} = [\bx'_1, \ldots, \bx'_{d_{\ell \ell m}} ] \in \mathbb{R}^{n \times d_{\ell \ell m}}
\end{equation}
where each row $f_{\ell\ell m}(d_i)=\bx_i$ is the embedding of document $d_i$.

\subsection{Leveraging Named Entities for Context-Aware Graph Construction}
We aim to train a similarity model, such as Word2Vec \cite{mikolov2013efficient}, to capture word-level similarities. Note that any other model can be used. The objective is to identify named entities, locate entities that occur in similar contexts using the trained model and cosine similarity, and build a document graph $\cal{\bar{G}}$ where edges represent similarities between named entities across documents.

Let ${\cal{D}} = \{d_1, d_2, \dots, d_n\}$ be the set of documents in the dataset. The problem is divided into the following steps:
\subsubsection{Step 1: Training a Similarity Model}
First, we train a Word2Vec model $f: {\cal{V}} \to \mathbb{R}^{e}$, where $\cal{V}$ is the vocabulary of the dataset, and each word $w_i \in \cal{V}$ is mapped to an embedding vector $f(w_i) \in \mathbb{R}^e$. The objective of Word2Vec is to maximize the likelihood of predicting the context words for a given target word $w_i$ using the following objective function:
\begin{equation}    
\max_{\theta} \sum_{w_i \in \cal{V}} \sum_{w_j \in \text{Context}(w_i)} \log P(w_j \mid w_i; \theta)
\end{equation}
where $\theta$ represents the parameters of the model and $P(w_j \mid w_i; \theta)$ is the probability that the word $w_j$ appears in the context of the word $w_i$.

\subsubsection{Step 2: Named Entity Recognition (NER)}
Let $\mathcal{E}_d = \{ e_1, e_2, \dots, e_{|E_d|} \}$ be the set of named entities extracted from a document $d \in \cal{D}$. We apply a NER model to each document $d_i$ to identify named entities $\mathcal{E}_{d_i}$. Formally, we define an NER model as:
$  
{\cal{NER}}: d  \to \mathcal{E}_d
$
where $\mathcal{E}_d$ is the set of named entities detected in document $d$.
\subsubsection{Step 3: Entity Similarity Search}
For each named entity $e \in \mathcal{E}_d$, we use the trained model $f$ to compute the cosine similarity between named entities in different documents. Given two entities $e_i$ and $e_j$, their similarity is defined as:
\begin{equation}    
\text{Sim}(e_i, e_j) = \frac{f(e_i) \cdot f(e_j)}{\|f(e_i)\| \|f(e_j)\|}.
\end{equation}
We identify pairs of named entities $(e_i, e_j)$ in documents that have a cosine similarity exceeding a threshold $\tau$.
\subsubsection{Step 4: Graph Construction (with Entity Threshold)}
We construct a document graph ${\cal{\bar{G}}} = (V_{\cal{\bar{G}}}, E_{\cal{\bar{G}}})$, where
$V_{\cal{\bar{G}}}$ represents the set of documents $\{d_1, \dots, d_n\}$ and $E_{\cal{\bar{G}}}$ represents the edges between documents. An edge exists between two documents $d_i$ and $d_j$ if they share a sufficient number of similar named entities. Specifically, an edge is formed only if the number of shared entities $\mathcal{E}_{d_i} \cap \mathcal{E}_{d_j}$ is greater than or equal to a predefined threshold $\tau$. The weight of the edge $a_{ij}$ is proportional to the similarity of the named entities between the two documents:
\begin{equation}    
a_{ij} = \frac{1}{|\mathcal{E}_{d_i} \cap \mathcal{E}_{d_j}|} \sum_{e_i \in \mathcal{E}_{d_i}, e_j \in \mathcal{E}_{d_j}} \text{Sim}(e_i, e_j)
\end{equation}
where $|\mathcal{E}_{d_i} \cap \mathcal{E}_{d_j}| \geq \tau$. This ensures that a link is established only when the documents share a sufficient number of similar entities, reducing the risk of connecting documents based on superficial relationships. The resulting graph is represented by an adjacency matrix $\bA_{ner}$, where $\bA_{ner}(i, j) = a_{ij}$ for documents $d_i$ and $d_j$ that satisfy the entity threshold $\tau$. The final objective is to harness this graph for document clustering.
\subsection{Joint Embedding and Clustering}
Our objective is to simultaneously learn node embeddings and cluster assignments. Inspired by \cite{fettal2022efficient}, we formulate the problem as follows:
\begin{equation}    
\begin{split}
\min_{\theta_1, \theta_2, \bG, \bF} 
\Big( &
\left\lVert \text{D}_{\theta_2} \left( \text{E}_{\theta_1} \left( \text{Agg}(\bA_{ner}, \bX_{\ell \ell m}) \right) \right) - \text{Agg}(\bA_{ner}, \bX_{\ell \ell m}) \right\rVert^2 \\
&+ \lambda \left\lVert \text{E}_{\theta_1} \left( \text{Agg}(\bA_{ner}, \bX_{\ell \ell m}) \right) - \bG \bF \right\rVert^2
\Big)
\end{split}
\end{equation}
\begin{equation*}   
\mbox{subject to: }
\bG \in \{0, 1\}^{n \times k}, \quad \bG \mathds{1}_k = \mathds{1}_n
\end{equation*}
where
$\text{E}_{\theta_1}$ and $\text{D}_{\theta_2}$ represent the encoding and decoding functions. The term \linebreak
$\text{Agg}(\bA_{ner}, \bX_{\ell \ell m})$ is an aggregation of the adjacency matrix $\bA_{ner}$ and node features $\bX_{\ell \ell m}$,
$\bG \in \{0, 1\}^{n \times k}$ is the binary cluster assignment matrix,
$\bF \in \mathbb{R}^{k \times d}$ represent the cluster centroids in the embedding space while the parameter
$\lambda$ controls the trade-off between reconstruction and clustering. Specifically,
the second term in the objective function can be viewed as optimized by {\tt Kmeans}. It is applied on the encoded representations forcing, thereby, the learned embeddings to be clustering-friendly. This penalizes representations that do not fit into clear clusters, following the loss of {\tt Kmeans}.
\paragraph{\bf Linear Graph Embedding Model.}
We use a linear graph autoencoder (LGAE) approach, which has been shown to perform comparably to more complex GCN-based models for tasks such as link prediction and node clustering \cite{kipf2016variational,salha2021simple}. Our encoder is defined as a simple linear transformation:
\begin{equation}    
\text{E}\left(\text{Agg}(\bA_{ner}, \bX_{\ell \ell m}); \bW_1\right) = \text{Agg}(\bA_{ner}, \bX_{\ell \ell m}) \bW_1
\end{equation}
In contrast to LGAE, which reconstructs the adjacency matrix $\bA_{ner}$ directly, the decoder incorporates both the adjacency matrix and the node features and is 
$ \bZ \bW_2$
where $\bZ = \text{Agg}(\bA_{ner}, \bX_{\ell \ell m}) \bW_1$, representing the encoded graph.

\paragraph{\bf Normalized Simple Graph Convolution.} Inspired by the Simple Graph Convolution (SGC) \cite{wu2019simplifying}, we define our aggregation function by
$  
\text{Agg}(\bA_{ner}, \bX_{\ell \ell m}) = \bT^p \bX_{\ell \ell m}
 $
where $\bT$ the symmetric normalized adjacency matrix with added self-loops is defined by
$
\bT = \bD_T^{-1}(\bI + \tilde{\bS})
$  
where $\tilde{\bS} = \tilde{\bD}^{-1/2} \tilde{\bA} \tilde{\bD}^{-1/2}$ with $\tilde{\bA} = \bA_{ner} + \bI$; $\tilde{\bD}$ and $\bD_T$ are the diagonal degree matrices of $\tilde{\bA}$ and $\bI+\tilde{\bS}$ respectively. This formulation extends traditional graph convolution by normalizing the spectrum of the graph filter, ensuring that the filter acts as a low-pass filter in the frequency range $[0, 1]$. In the following, for convenience, denote the matrix $\bT^p \bX_{\ell\ell m}$ by $\bY^p$.
\paragraph{\bf Optimization Problem.}
The objective function takes the following form:
\begin{equation}\label{eq1}    
\min_{\bG, \bF, \bW_1, \bW_2} \left\lVert \bY^p - \bY^p \bW_1 \bW_2 \right\rVert^2 + \lambda \left\lVert \bY^p \bW_1 - \bG \bF \right\rVert^2
\end{equation}
\begin{equation*} 
\mbox{subject to: }
\bG \in \{0, 1\}^{n \times k}, \quad \bG \mathds{1}_k = \mathds{1}_n
\end{equation*}
The two terms in (\ref{eq1}) establish a link between the two tasks, with the first term acting as a linear autoencoder and the second term facilitating clustering in the embedding space. The parameter $\lambda$ controls the importance of the second term in terms of regularizing the embedding. However, we take $\lambda=1$ as in \cite{fettal2022efficient}; this assumption can be investigated in the future even it appears effective in our experiments.
\paragraph{\bf Graph Convolutional Clustering.} To further enhance the interaction between embedding and clustering, we assume $\bW = \bW_1 = \bW_2^\top$ and impose an orthogonality constraint $\bW^\top \bW = \bI_k$, leading to the modified problem:
\begin{equation}    
\min_{\bG,\bF,\bW} \left\lVert \bY^p - \bY^p \bW \bW^\top \right\rVert^2 + \left\lVert \bY^p \bW - \bG \bF \right\rVert^2
\end{equation}
\begin{equation*}  
\mbox{subject to: }
\bG \in \{0, 1\}^{n \times k}, \quad \bG \mathds{1}_k = \mathds{1}_n, \quad \bW^\top \bW = \bI_k
\end{equation*}
This formulation encourages a more direct interaction between the embedding and clustering tasks. Following \cite{allab2016semi,allab2018simultaneous,labiod2021efficient,fettal2022efficient}, we can show that solving this problem is equivalent to solving (subject to the same constraints):
\begin{equation} \label{eq7}
\min_{\bG,\bF,\bW} \left\lVert \bY^p- \bG \bF \bW^\top \right\rVert^2.
\end{equation}
By decomposing the reconstruction and regularization terms, it can be shown that both formulations lead to the same solution, allowing efficient joint learning of embeddings and clusters. The solution of the classical problem (\ref{eq7}) is accomplished by alternating updates of $\bG$, $\bF$ and $\bW$; all steps are detailed in \cite{fettal2022efficient} where $\bW$ is initialized using a randomized PCA on $\bY^p$ and $\bG$ is initialized using {\tt Kmeans} on $\bY^p\bW$.

%
%
\begin{algorithm}
\footnotesize
\caption{{\tt GCC$^*$}: GCC incorporating named entities and LLM}
\begin{algorithmic}
\REQUIRE Dataset $\cal{D}$, $f_{\ell \ell m}$, similarity threshold $\tau$, number of clusters $k$
\ENSURE Clustered documents
\STATE{$\leadsto$} Extract named entities $\mathcal{E}_d$ for each $d \in \cal{D}$
\STATE{$\leadsto$} Train Word2Vec on $\cal{D}$
\STATE{$\leadsto$} Initialize list $\mathcal{L}$ to store document pairs
\FOR{each pair of docs $(d_i, d_j)$}
    \STATE{$\leadsto$} Compute similarity between $\mathcal{E}_{d_i}$ and $\mathcal{E}_{d_j}$
    \IF{similarity $> \tau$}
        \STATE{$\leadsto$} Append $(d_i, d_j)$ to $\mathcal{L}$
    \ENDIF
\ENDFOR
\FOR{each pair $(d_i, d_j)$ in $\mathcal{L}$}
    \IF{$d_i$ and $d_j$ share \textbf{at least 3 common named entities}}
        \STATE{$\leadsto$} Add edge between $d_i$ and $d_j$
    \ENDIF
\ENDFOR
\STATE{$\leadsto$} Construct graph $\bA_{ner}$ with docs as nodes and similarities as edges
\STATE{$\leadsto$} Generate embeddings $\bX_{\ell\ell m}$ for each doc using $f_{\ell \ell m}$
\STATE{$\leadsto$} Apply {\tt GCC} \cite{fettal2022efficient} on $(\bA_{ner}, \bX_{\ell\ell m})$ and obtain optimal $\bG$, $\bF$ and $\bW$
\STATE{$\leadsto$} Deduce the clusters of documents from $\bG$.
\end{algorithmic}
\end{algorithm}
\section{Experiments}
\subsection{Datasets}
We assess our model using different configurations: $\bX_{co}$ (Bag-of-Words), $\bX_{\ell\ell m}$ (LLM embeddings), $\bA_{knn}$ (KNN-based graph), and $\bA_{ner}$ (NER-based graph). We test our model using four datasets that vary in size and number of clusters. The characteristics of these datasets are shown in Table \ref{datasets}.
\begin{table}[!ht]
\caption{Description of datasets. The balance represents the ratio between the smallest and largest class. \#Tokens indicates the mean token count.}
\footnotesize
\centering
\begin{tabular}{ccccccc}
\hline
Datasets & \multicolumn{6}{c}{Characteristics}  \\
\cline{2-7}
 & \#Documents & \#Clusters & Balance & \#Tokens & Domaine & Language\\
\hline
BBC News\footnote{http://mlg.ucd.ie/datasets/bbc.html} & 2,225 & 5 & 0.75 & 390 & News articles & En\\
MLSUM \cite{scialom2020mlsum} & 407,835 & 612 & 2.2e-05 & 543 & News articles & Fr\\
Arxiv-10 \cite{farhangi2022protoformer} & 100,000 & 10 & 1 & 155 & Scientific papers & En\\
PubMed & 19,716 & 3 & 0.52 & 224 & Pubmed abstracts & En\\
\hline
\end{tabular}
\label{datasets}
\end{table}
\subsection{Clustering algorithms}
In this section, we evaluate various clustering algorithms using different input types: the document feature matrix $\bX{.}$, the adjacency matrix $\bA{.}$, or a combination of both $(\bX_{.}, \bA_{.})$. These matrices capture different aspects of the data:
\begin{itemize}
    \item {\tt Kmeans} \cite{macqueen1967some}, {\tt Deep Kmeans} \cite{fard2020deep} and {\tt DCN} \cite{yang2017towards} 
    : Use $\bX_{co}$ or $\bX_{\ell\ell m}$.
    \item {\tt Spectral clustering}: Uses only $\bA_{ner}$ or $\bA_{knn}$ for clustering.
    \item {\tt PC Kmeans}: Uses $(\bX_{\ell\ell m}, \bA_{ner})$ or $(\bX_{\ell\ell m}, \bA_{knn})$. We retain the top 500 strongest links from the named entity graph $\bA_{*}$ as must-link constraints. Increasing this number reduced model performance, so we enforced the 500-link limit.  
    \item {\tt GCC$_{(.,.)}$}: applied to  $(\bA_{knn},\bX_{co})$, $(\bA_{ner},\bX_{co})$ and $(\bA_{ner},\bX_{\ell\ell m})$ respectively.
    \item {\tt GCC$^*$}: applied to $(\bA_{ner}, \bX_{\ell\ell m})$.
\end{itemize}
%
\subsection{Experimental setting}
Using labeled datasets, we evaluate clustering algorithms performance with external indices: Accuracy (ACC), Normalized Mutual Information (NMI) \cite{strehl2002cluster}, and Adjusted Rand Index (ARI) \cite{steinley2004properties}. ACC measures how well each cluster matches the ground-truth class labels, while NMI, ranging from 0 to 1, evaluates the mutual dependence between the predicted clusters and true labels. Finally, the ARI considers the similarity between predicted clusters and ground truth partitions, with a range from -0.5 to 1, where higher values indicate better agreement. Intuitively, NMI quantifies how much the estimated clustering is informative about the true clustering, while the ARI measures the degree of agreement between the estimated clustering and the reference partition. Both NMI and ARI are equal to 1 if the resulting clustering partition is identical to the ground truth. Contrary to ACC, it is important to note that NMI and ARI are more reliable as external indices because they are less sensitive to disproportionate classes.

For clustering algorithms, {\tt Kmeans} is initialized using {\tt K-means++} \cite{arthur2006k}, which selects starting centroids by sampling based on their contribution to total inertia. We limit the iterations to 300 and run 10 initial setups to enhance clustering stability. For {\tt Deep Kmeans} and {\tt DCN}, we use the default settings to maintain consistency across experiments.

We remove stopwords and limit features to 2000 for co-occurrence matrices. LLM embeddings were generated using OpenAI's \emph{text-embedding-3-small} model (1536 dimensions, \$0.02 per 1M tokens). Documents exceeding 8,191 tokens were excluded to fit model limits. Due to computational constraints, we randomly sample 10,000 Arxiv documents and 16,321 MLSUM documents from five categories: Sport, Health, Politics, Economy, and Climate.

For the NER, we use \textit{camembert-ner}\footnote{https://huggingface.co/Jean-Baptiste/camembert-ner} model for MLSUM since the dataset is in French, and DeBERTa \cite{ushio-camacho-collados-2021-ner} for BBC News. For other datasets, we used \textit{GPT-4o}\footnote{https://platform.openai.com/docs/models/gpt-4o} with prompts to extract named entities in JSON format, where keys are entity types and values are lists of entities. For the named entity graph, we only kept links where $\tau$ was greater than $0.9$ between each pair of named entities of the same type. A link between two documents was considered only if there were at least three links between their named entities. Word2Vec was trained using CBOW with \texttt{num\_features = 500}, \texttt{min\_word\_count = 10}, context window size of \texttt{5}, and 20 epochs. Composite named entities were merged to be treated as a single token by Word2Vec.

\subsection{Assessing the number of clusters $k$ and the power $p$}
\begin{figure}[ht!]
  \centering
\fbox{\includegraphics[width=13cm]{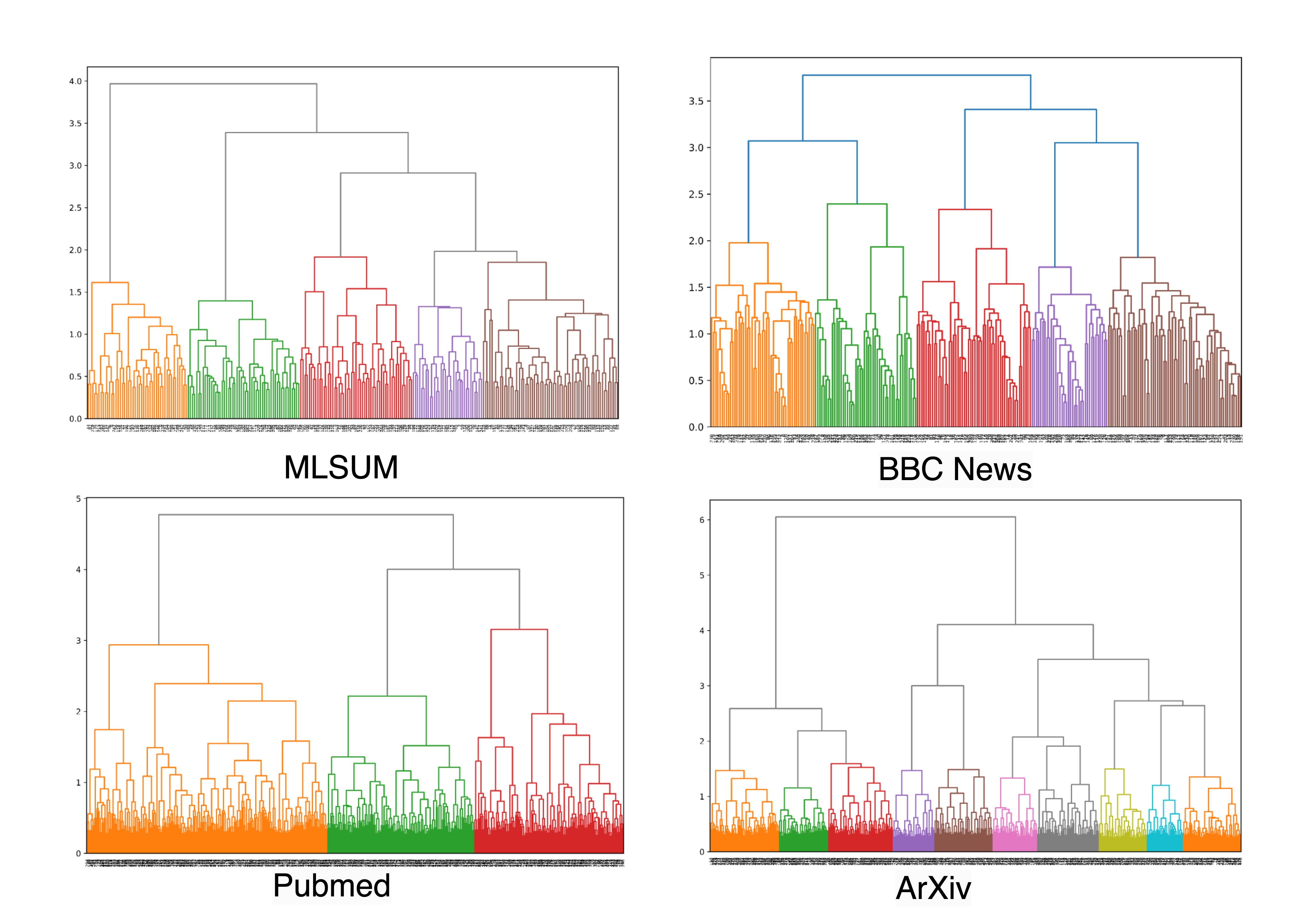}}
  \caption{Comparison of dendrograms for different datasets.}
  \label{dendrogramme}
\end{figure}

Optimizing the propagation parameter $p$ and the number of clusters $k$ is essential for the performance of our method. The parameter $p$ controls the neighborhood information captured by the GCN and affects the smoothness of node embeddings. An appropriate $p$ aggregates enough information without over-smoothing. Similarly, choosing the right number of clusters $k$ ensures proper grouping of documents, avoiding over- or under-segmentation.
\paragraph{\bf Number of clusters $k$.} Since our framework is unsupervised, we aim to assess whether we could accurately detect the true number of clusters. Traditional internal criteria such as silhouette score \cite{rousseeuw1987silhouettes}, Davies-Bouldin index \cite{davies1979cluster}, and Calinski-Harabasz index \cite{calinski1974dendrite} did not yield satisfactory results. Most of these metrics tend to favor the minimum number of clusters specified in the grid search. To address this limitation, we applied an alternative approach by running {\tt GCC$^*$} with a large number of clusters - 250 for BBC News and 500 for other datasets. This over-segmentation allows us to capture local patterns, where each cluster centroid represents a dense region in the feature space. We then applied hierarchical clustering using Ward's method \cite{ward1963hierarchical} on the centroids. This method progressively merges clusters, and by observing the dendrogram (Figure \ref{dendrogramme}), we selected the appropriate number of clusters. For most datasets, the true number of clusters was easily identified. For the arXiv dataset, several partitions can be considered, with 3, 4, or 10 clusters. To remain consistent with the benchmarks and our evaluation study of {\tt GCC$^*$}, we opted for 10 clusters.
\paragraph{\bf Power $p$.}
Once the number of classes has been set, our objective is to estimate $p$. We tested a range of values for $p$ between 1 and 50 and selected the optimal value by minimizing the square root of the cluster loss. This ensures that the chosen $p$ captures sufficient neighborhood information while avoiding oversmoothing of the graph signal. We observed that a value of $p=2$ or $p=3$ is the value retained for the 4 datasets; the difference between these two values is not significant. 
\subsection{Results}
\paragraph{\bf Quality of clustering.} In Table \ref{results} we compare the performance of different clustering algorithms on four datasets: BBC News, MLSUM, PubMed, and Arxiv-10. We observe that the algorithms using co-occurrence matrices perform significantly worse compared to those using LLM embeddings. For example, on the BBC News dataset, \texttt{Deep Kmeans} with $\bX_{co}$ achieves the highest accuracy (66.45\%), but its performance on larger datasets such as MLSUM and PubMed drops significantly (NMI < 8\% and ARI < 2\%). This reflects the inability of co-occurrence-based representations to capture deeper semantic relationships. On the other hand, the methods incorporating LLM embeddings show a substantial improvement in clustering performance. For example, on the BBC News dataset, \texttt{Kmeans} achieves 93.01\% ACC and 91.01 ARI, while \texttt{DCN} reaches 86.76\% ACC. This improvement is consistent across datasets, highlighting the advantage of using rich contextual embeddings.

The proposed approach ({\tt GCC$^*$}) on $(\bA_{ner},\bX_{\ell \ell m})$, exceeds all other methods, including KNN graph-based clustering. In the BBC News dataset, {\tt GCC$^*$}  achieves a remarkable NMI of 95.12\%, demonstrating that the incorporation of the similarities of the named entities into the graph structure significantly improves clustering performance. A similar pattern is seen in other datasets, such as MLSUM, where {\tt GCC$^*$}  achieves an NMI of 72.42\%. The results of spectral clustering using different adjacency matrices $\bA$ show that simply using $\bA_{knn}$ or $\bA_{ner}$ is not sufficient to achieve good clustering performance. The main challenge lies in the structure of these adjacency matrices, which fail to capture meaningful relationships for clustering in certain scenarios. 
\begin{table}
\centering
\small
\caption{Clustering performance on four datasets averaged over 20 runs}
\setlength{\tabcolsep}{4.6pt}
\begin{tabularx}{\textwidth}{l|c|ccc|ccc|ccc|ccc}
\hline
\textbf{Method} & \textbf{Input} & \multicolumn{3}{c|}{\textbf{BBC}} & \multicolumn{3}{c|}{\textbf{MLSUM}} & \multicolumn{3}{c|}{\textbf{PubMed}} & \multicolumn{3}{c}{\textbf{ArXiv-10}} \\
 &  & ACC & NMI & ARI & ACC & NMI & ARI & ACC & NMI & ARI & ACC & NMI & ARI \\ \hline \hline
 {\tt Kmeans}& $\bX_{co}$ & 43.91 & 27.16 & 13.47 & 24.86 & 0.60 & 0.40 & 44.16 & 14.12 & 8.77 & 37.54 & 29.80 & 12.30 \\
{\tt DKmeans} & $\bX_{co}$ & 66.45 & 56.11 & 47.73 & 25.63 & 0.61 & 0.42 & 44.83 & 7.86 & 1.89 & 41.23 & 31.19 & 16.67 \\
{\tt DCN} & $\bX_{co}$ & 58.58 & 40.49 & 25.95 & 26.87 & 2.01  & 0.99 & 45.69 & 15.41 & 8.45 & 35.85 & 27.56 & 12.63 \\ 

\hline
{\tt Kmeans}& $\bX_{\ell \ell m}$ & 93.01 & 87.34 & 91.01 & 75.62 & 66.28 & 62.60 & 63.25 & 27.02 & 23.58 & 69.08 & 60.11 & 51.74 \\
{\tt DKmeans} & $\bX_{\ell \ell m}$ & 85.51 & 75.10 & 73.62 & 75.01 & 58.25 & 57.41 & 55.54 & 17.85 & 13.90 & 69.26 & 59.62 & 51.71 \\
{\tt DCN} & $\bX_{\ell \ell m}$ & 86.76 & 72.78 & 71.85 & 70.14 & 52.99 & 49.04 & 58.71 & 25.44 & 18.32 & 66.45 & 56.11 & 47.73 \\ 

\hline
{\tt Spectral} & $\bA_{knn,\ell\ell m}$ & 34.11 & 19.32 & 3.82 & 31.90 & 0.9 & 0.10 & 40.01 & 0.08 & 0.06 & 14.86 & 5.32 & 0.6 \\
{\tt Spectral} & $\bA_{ner,\ell\ell m}$ & 35.01 & 20.01 & 7.87 & 31.33 & 1.23 & 0.20 & 39.93 & 0.20 & 0.0 & 20.01 & 6.31 & 2.12 \\ \hline
{\tt PCKmeans} & $\bA_{knn},\bX_{\ell\ell m}$ & 70.85 & 73.00 & 63.89 & 70.01 & 49.00 & 45.80 & 60.01 & 24.24 & 22.87 & 68.12 & 60.90 & 52.14 \\
{\tt PCKmeans} & $\bA_{ner},\bX_{\ell\ell m}$ & 70.88 & 73.06 & 64.01 & 70.71 & 49.87 & 45.88 & 60.21 & 24.89 & 23.06 & 69.01 & 61.90 & 52.34 \\
\hline \hline
{\tt GCC$_{({knn},{co})}$} & $\bA_{knn},\bX_{co}$ & 88.76 & 76.26 & 75.97 & 25.20 & 0.60 & 0.71 & 63.02 & 25.20 & 24.78 & 60.12 & 47.75 & 39.08\\
{\tt GCC$_{({ner},{co})}$} & $\bA_{ner},\bX_{co}$ & 95.43 & 86.18 & 89.27 & 27.69 & 2.91 & 3.80 & 63.00 & 25.23 & 24.80 & 60.01 & 49.57 & 40.01 \\
\hline
{\tt GCC$_{({knn},{\ell\ell m})}$} & $\bA_{knn},\bX_{\ell\ell m}$ & 95.82 & 87.41 & 90.12 & 76.56 & 67.12 & 64.99 & 65.00 & 28.34 & \textbf{25.00} & 70.33 & 60.70 & 52.01 \\ 
{\tt GCC$^*$} & $\bA_{ner},\bX_{\ell\ell m}$ & \textbf{97.61} & \textbf{95.12} & \textbf{96.66} & \textbf{88.32} & \textbf{72.42} & \textbf{74.58} & \textbf{65.10} & \textbf{29.04} & \textbf{25.00} & \textbf{73.2} & \textbf{63.98} & \textbf{55.8} \\ \hline
\end{tabularx}
\label{results}
\end{table}
Indeed, the use of $\bA_{knn}$ often results in too many links between nodes, leading to a densely connected graph. This excessive connectivity diminishes the quality of the clusters as many unrelated points are grouped together. For example, on the BBC dataset, spectral clustering with $\bA_{knn,\ell\ell m}$ achieves an ARI of only 3.82\%, highlighting the negative impact of overly dense connections. However, although $\bA_{ner}$ captures entity-level information, it tends to produce a sparse graph, especially in large and diverse documents. This sparseness leads to weakly connected clusters, as seen in the PubMed dataset, where spectral clustering with $\bA_{ner}$ yields an NMI close to zero. The poor performance of spectral clustering emphasizes the need for more balanced graph structures that reflect both semantic and entity-level similarities.
\paragraph{\bf Quality of embedding.}
We evaluate the quality of the embedding through the visualization of the truth classes using UMAP (default parameters) based on $\bX_{\ell\ell m} \in \mathds{R}^{n\times d}$ and $\bY^p\bW \in \mathds{R}^{n\times k}$ which is derived from {\tt GCC$^*$}. It is remarkable to observe the quality of class separability that we illustrate in Figure \ref{fig.graph.ner}.
\begin{figure}[!ht] 
\centering \includegraphics[width=10cm]{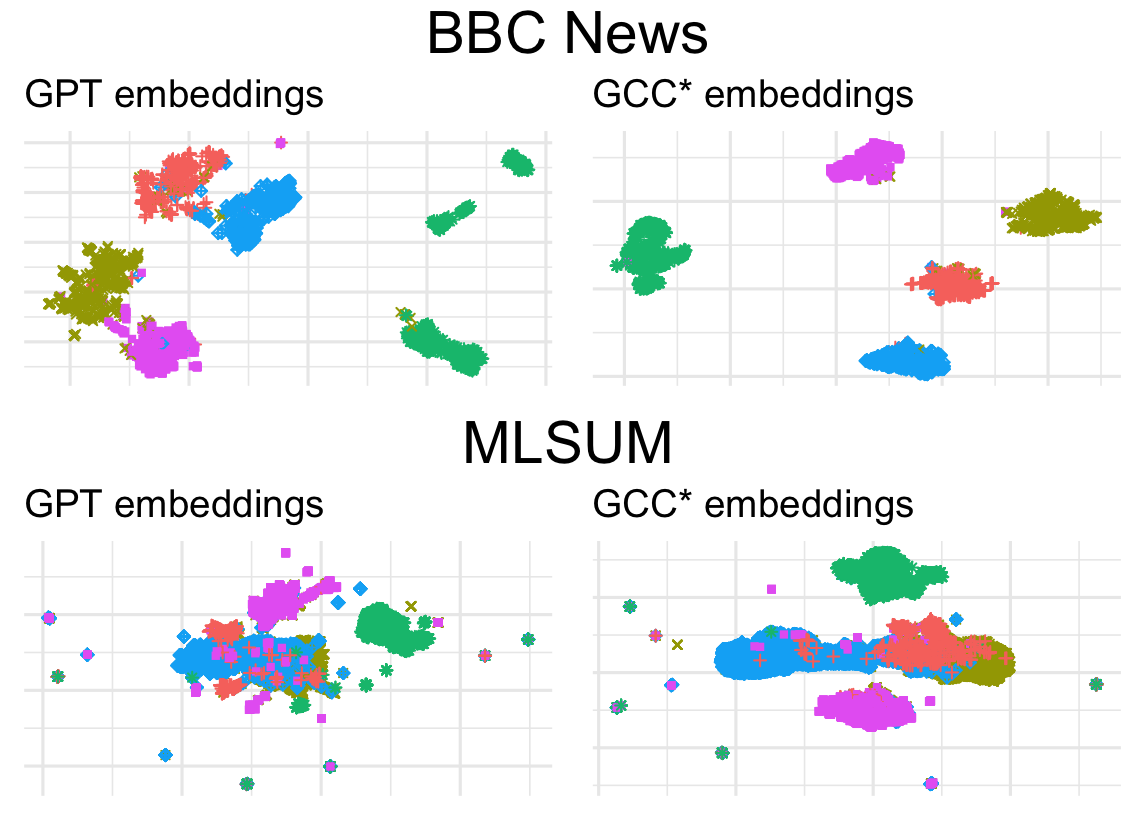} 
\caption{UMAP projection of the cluster embeddings obtained with GPT ($\bX_{\ell \ell m}$) compared to those obtained on $\bY^p\bW$ derived from {\tt GCC$^*$}.} 
\label{fig.graph.ner} 
\end{figure}
\section{CONCLUSION}
This study introduces an innovative method for document clustering that integrates NER and LLM embeddings in the attributed graphs framework. By building a document graph using entity similarities and incorporating contextual embeddings, our approach outperformed traditional clustering techniques. The application of GCN facilitates the simultaneous optimization of embeddings and clustering, leading to more effective grouping of semantically similar documents.

Future work could focus on exploring whether specific types of named entities (e.g., organizations, locations, or events) are more relevant than others for certain clustering tasks. This direction could provide more refined document groupings and reveal the underlying thematic relationships driven by particular entity types.

\bibliographystyle{acm}
\bibliography{biblio}

\begin{thebibliography}{10}

\bibitem{affeldt2021regularized}
{\sc Affeldt, S., Labiod, L., and Nadif, M.}
\newblock Regularized bi-directional co-clustering.
\newblock {\em Statistics and Computing 31}, 3 (2021), 32.

\bibitem{allab2016semi}
{\sc Allab, K., Labiod, L., and Nadif, M.}
\newblock A semi-nmf-pca unified framework for data clustering.
\newblock {\em IEEE TKDE 29}, 1 (2016), 2--16.

\bibitem{allab2018simultaneous}
{\sc Allab, K., Labiod, L., and Nadif, M.}
\newblock Simultaneous spectral data embedding and clustering.
\newblock {\em IEEE TNNLS 29}, 12 (2018), 6396--6401.

\bibitem{arthur2006k}
{\sc Arthur, D., and Vassilvitskii, S.}
\newblock k-means++: The advantages of careful seeding.
\newblock Tech. rep., Stanford, 2006.

\bibitem{calinski1974dendrite}
{\sc Cali{\'n}ski, T., and Harabasz, J.}
\newblock A dendrite method for cluster analysis.
\newblock {\em Communications in Statistics-theory and Methods 3}, 1 (1974), 1--27.

\bibitem{cao2012text}
{\sc Cao, T.~H., Tang, T.~M., and Chau, C.~K.}
\newblock Text clustering with named entities: a model, experimentation and realization.
\newblock In {\em Data Mining: Foundations and Intelligent Paradigms}. 2012, pp.~267--287.

\bibitem{davies1979cluster}
{\sc Davies, D.~L., and Bouldin, D.~W.}
\newblock A cluster separation measure.
\newblock {\em IEEE transactions on pattern analysis and machine intelligence}, 2 (1979), 224--227.

\bibitem{derczynski2015analysis}
{\sc Derczynski, L., Maynard, D., Rizzo, G., Van~Erp, M., Gorrell, G., Troncy, R., Petrak, J., and Bontcheva, K.}
\newblock Analysis of named entity recognition and linking for tweets.
\newblock {\em Information Processing \& Management 51}, 2 (2015), 32--49.

\bibitem{fard2020deep}
{\sc Fard, M.~M., Thonet, T., and Gaussier, E.}
\newblock Deep k-means: Jointly clustering with k-means and learning representations.
\newblock {\em Pattern Recognition Letters 138\/} (2020), 185--192.

\bibitem{farhangi2022protoformer}
{\sc Farhangi, A., Sui, N., Hua, N., Bai, H., Huang, A., and Guo, Z.}
\newblock Protoformer: Embedding prototypes for transformers.
\newblock In {\em PAKDD\/} (2022), pp.~447--458.

\bibitem{fettal2022efficient}
{\sc Fettal, C., Labiod, L., and Nadif, M.}
\newblock Efficient graph convolution for joint node representation learning and clustering.
\newblock In {\em WSDM\/} (2022), pp.~289--297.

\bibitem{hassani2020contrastive}
{\sc Hassani, K., and Khasahmadi, A.~H.}
\newblock Contrastive multi-view representation learning on graphs.
\newblock In {\em ICML\/} (2020), pp.~4116--4126.

\bibitem{keraghel2024beyond}
{\sc Keraghel, I., Morbieu, S., and Nadif, M.}
\newblock Beyond words: a comparative analysis of llm embeddings for effective clustering.
\newblock In {\em IDA\/} (2024), pp.~205--216.

\bibitem{kim2021knn}
{\sc Kim, J.-H., Choi, J.-H., Park, Y.-H., Leung, C. K.-S., and Nasridinov, A.}
\newblock Knn-sc: novel spectral clustering algorithm using k-nearest neighbors.
\newblock {\em IEEE Access 9\/} (2021), 152616--152627.

\bibitem{kipf2016semi}
{\sc Kipf, T.~N., and Welling, M.}
\newblock Semi-supervised classification with graph convolutional networks.
\newblock {\em arXiv preprint arXiv:1609.02907\/} (2016).

\bibitem{kipf2016variational}
{\sc Kipf, T.~N., and Welling, M.}
\newblock Variational graph auto-encoders.
\newblock {\em arXiv preprint arXiv:1611.07308\/} (2016).

\bibitem{labiod2021efficient}
{\sc Labiod, L., and Nadif, M.}
\newblock Efficient regularized spectral data embedding.
\newblock {\em Advances in Data Analysis and Classification 15}, 1 (2021), 99--119.

\bibitem{lee2018pre}
{\sc Lee, J., and Toutanova, K.}
\newblock Pre-training of deep bidirectional transformers for language understanding.
\newblock {\em arXiv preprint arXiv:1810.04805 3}, 8 (2018).

\bibitem{macqueen1967some}
{\sc Macqueen, J.}
\newblock Some methods for classification and analysis of multivariate observations.
\newblock In {\em Proceedings of 5-th Berkeley Symposium on Mathematical Statistics and Probability/University of California Press\/} (1967).

\bibitem{mikolov2013efficient}
{\sc Mikolov, T.}
\newblock Efficient estimation of word representations in vector space.
\newblock {\em arXiv preprint arXiv:1301.3781\/} (2013).

\bibitem{muennighoff2022mteb}
{\sc Muennighoff, N., Tazi, N., Magne, L., and Reimers, N.}
\newblock Mteb: Massive text embedding benchmark.
\newblock {\em arXiv preprint arXiv:2210.07316\/} (2022).

\bibitem{ng2001spectral}
{\sc Ng, A., Jordan, M., and Weiss, Y.}
\newblock On spectral clustering: Analysis and an algorithm.
\newblock {\em Advances in neural information processing systems 14\/} (2001).

\bibitem{qin2018novel}
{\sc Qin, Y., Yu, Z.~L., Wang, C.-D., Gu, Z., and Li, Y.}
\newblock A novel clustering method based on hybrid k-nearest-neighbor graph.
\newblock {\em Pattern recognition 74\/} (2018), 1--14.

\bibitem{radford2018improving}
{\sc Radford, A.}
\newblock Improving language understanding by generative pre-training.

\bibitem{reimers2019sentence}
{\sc Reimers, N.}
\newblock Sentence-bert: Sentence embeddings using siamese bert-networks.
\newblock {\em arXiv preprint arXiv:1908.10084\/} (2019).

\bibitem{rousseeuw1987silhouettes}
{\sc Rousseeuw, P.~J.}
\newblock Silhouettes: a graphical aid to the interpretation and validation of cluster analysis.
\newblock {\em Journal of computational and applied mathematics 20\/} (1987), 53--65.

\bibitem{salah2019directional}
{\sc Salah, A., and Nadif, M.}
\newblock Directional co-clustering.
\newblock {\em Advances in Data Analysis and Classification 13\/} (2019), 591--620.

\bibitem{salha2021simple}
{\sc Salha, G., Hennequin, R., and Vazirgiannis, M.}
\newblock Simple and effective graph autoencoders with one-hop linear models.
\newblock In {\em ECML-PKDD\/} (2021), pp.~319--334.

\bibitem{schutze2008introduction}
{\sc Sch{\"u}tze, H., Manning, C.~D., and Raghavan, P.}
\newblock {\em Introduction to information retrieval}, vol.~39.
\newblock Cambridge University Press Cambridge, 2008.

\bibitem{scialom2020mlsum}
{\sc Scialom, T., Dray, P.-A., Lamprier, S., Piwowarski, B., and Staiano, J.}
\newblock Mlsum: The multilingual summarization corpus.
\newblock {\em arXiv preprint arXiv:2004.14900\/} (2020).

\bibitem{sia2020tired}
{\sc Sia, S., Dalmia, A., and Mielke, S.~J.}
\newblock Tired of topic models? clusters of pretrained word embeddings make for fast and good topics too!
\newblock {\em arXiv preprint arXiv:2004.14914\/} (2020).

\bibitem{steinley2004properties}
{\sc Steinley, D.}
\newblock Properties of the hubert-arable adjusted rand index.
\newblock {\em Psychological methods 9}, 3 (2004), 386.

\bibitem{strehl2002cluster}
{\sc Strehl, A., and Ghosh, J.}
\newblock Cluster ensembles---a knowledge reuse framework for combining multiple partitions.
\newblock {\em JMLR 3\/} (2002), 583--617.

\bibitem{ushio-camacho-collados-2021-ner}
{\sc Ushio, A., and Camacho-Collados, J.}
\newblock {T}-{NER}: An all-round python library for transformer-based named entity recognition.
\newblock In {\em EACL\/} (2021), pp.~53--62.

\bibitem{vaswani2017attention}
{\sc Vaswani, A.}
\newblock Attention is all you need.
\newblock {\em Advances in Neural Information Processing Systems\/} (2017).

\bibitem{velickovic2019deep}
{\sc Velickovic, P., Fedus, W., Hamilton, W.~L., Li{\`o}, P., Bengio, Y., and Hjelm, R.~D.}
\newblock Deep graph infomax.
\newblock {\em ICLR (Poster) 2}, 3 (2019), 4.

\bibitem{wang2016structural}
{\sc Wang, D., Cui, P., and Zhu, W.}
\newblock Structural deep network embedding.
\newblock In {\em SIGKDD\/} (2016), pp.~1225--1234.

\bibitem{ward1963hierarchical}
{\sc Ward~Jr, J.~H.}
\newblock Hierarchical grouping to optimize an objective function.
\newblock {\em Journal of the American statistical association 58}, 301 (1963), 236--244.

\bibitem{wei2015semantic}
{\sc Wei, T., Lu, Y., Chang, H., Zhou, Q., and Bao, X.}
\newblock A semantic approach for text clustering using wordnet and lexical chains.
\newblock {\em Expert Systems with applications 42}, 4 (2015), 2264--2275.

\bibitem{wu2019simplifying}
{\sc Wu, F., Souza, A., Zhang, T., Fifty, C., Yu, T., and Weinberger, K.}
\newblock Simplifying graph convolutional networks.
\newblock In {\em ICML\/} (2019), pp.~6861--6871.

\bibitem{yang2017towards}
{\sc Yang, B., Fu, X., Sidiropoulos, N.~D., and Hong, M.}
\newblock Towards k-means-friendly spaces: Simultaneous deep learning and clustering.
\newblock In {\em ICML\/} (2017), pp.~3861--3870.

\bibitem{you2020graph}
{\sc You, Y., Chen, T., Sui, Y., Chen, T., Wang, Z., and Shen, Y.}
\newblock Graph contrastive learning with augmentations.
\newblock {\em Advances in neural information processing systems 33\/} (2020), 5812--5823.

\bibitem{zhang2020graph}
{\sc Zhang, J., Zhang, H., Xia, C., and Sun, L.}
\newblock Graph-bert: Only attention is needed for learning graph representations.
\newblock {\em arXiv preprint arXiv:2001.05140\/} (2020).

\end{thebibliography}

\end{document}